\documentclass{article}

\usepackage{microtype}
\usepackage{graphicx}
\usepackage{subfigure}
\usepackage{booktabs} % for professional tables

\usepackage[table]{xcolor}

\usepackage[accepted]{icml2025}

\usepackage{latexsym}
\usepackage{graphicx}
\usepackage{mathptmx}
\usepackage{nameref}

\usepackage{amsmath}
\usepackage{amsfonts}
\usepackage{amssymb}
\usepackage{amsbsy}
\usepackage{amsthm}
\usepackage{times}
\usepackage{epsfig}
\usepackage{graphicx}
\usepackage{amsmath}
\usepackage{amssymb}
\usepackage{booktabs}
\usepackage{soul}
\usepackage{multirow}
\usepackage{todonotes}
\usepackage{listings}
\usepackage[pdftex,colorlinks=true,urlcolor=blue,citecolor=black,anchorcolor=black,linkcolor=black,bookmarks=false]{hyperref}

\usepackage{hyphenat}
\newtheoremstyle{scsthe}% hnamei
{8pt}% hSpace abovei
{8pt}% hSpace belowi
{\it}% hBody fonti
{}% hIndent amounti1
{\bf}% hTheorem head fontbf
{.}% hPunctuation after theorem headi
{.5em}% hSpace after theorem headi2
{}% hTheorem head spec (can be left empty, meaning `normal')i

\theoremstyle{scsthe}

\begin{document}
\twocolumn[
\icmltitle{A Framework for Adversarial Analysis of Decision Support Systems Prior to Deployment}



\icmlsetsymbol{equal}{0}
% AUTHOR: Enter the authors of the article

\begin{icmlauthorlist}
\icmlauthor{Brett Bissey}{equal,yyy}
\icmlauthor{Kyle Gatesman}{equal,yyy}
\icmlauthor{Walker Dimon}{yyy}
\icmlauthor{Mohammad Alam}{yyy}
\icmlauthor{Luis Robaina}{yyy}
\icmlauthor{Joseph Weissman}{yyy}

\icmlaffiliation{yyy}{AI & Autonomy Center, MITRE Labs, McLean, VA, United States}

\end{icmlauthorlist}

\icmlcorrespondingauthor{Brett Bissey}{bbissey@mitre.org}
\icmlcorrespondingauthor{Kyle Gatesman}{kjgatesman@mitre.org}

\icmlkeywords{Reinforcement Learning, Model Evaluation, Adversarial Attacks, Explainable AI, Simulation Assurance}

\vskip 0.3in
]

%\footnote{*Equal Contribution}
\footnotetext[0\def\thefoornote{}]{Equal Contribution}
\footnotetext[1\def\thefoornote{}]{AI \& Autonomy Center, MITRE Labs, McLean, VA, United States. Correspondence to: Brett Bissey (bbissey@mitre.org), Kyle Gatesman (kjgatesman@mitre.org) \newline Copyright © 2024 The MITRE
Corporation. ALL RIGHTS RESERVED. Approved for Public Release; Distribution Unlimited. Public Release
Case Number 24-2499}

\begin{abstract}
This paper introduces a comprehensive framework designed to analyze and secure decision-support systems trained with Deep Reinforcement Learning (DRL), prior to deployment, by providing insights into learned behavior patterns and vulnerabilities discovered through simulation. The introduced framework aids in the development of precisely timed and targeted observation perturbations, enabling researchers to assess adversarial attack outcomes within a strategic decision-making context. We validate our framework, visualize agent behavior, and evaluate adversarial outcomes within the context of a custom-built strategic game, CyberStrike. Utilizing the proposed framework, we introduce a method for systematically discovering and ranking the impact of attacks on various observation indices and time-steps, and we conduct experiments to evaluate the transferability of adversarial attacks across agent architectures and DRL training algorithms. The findings underscore the critical need for robust adversarial defense mechanisms to protect decision-making policies in high-stakes environments.
\end{abstract}

\section{Introduction}
AI-enabled decision support systems trained in simulation are increasingly being deployed in safety-critical environments, making them vulnerable targets to adversarial attacks. Deep reinforcement learning (DRL) has been effective in training superhuman policies in strategic board games \cite{muzero}, video games like StarCraft \cite{alphastar}, robotics tasks \cite{roboticsdrl}, and autonomous driving \cite{kiran2021deepreinforcementlearningautonomous}. However due to the reliance on deep neural networks (DNN) for decision-making, analyzing the strengths and vulnerabilities of DNN policies trained with DRL requires additional methodology. Adversarial attacks can manipulate the system's perception of the environment through difficult-to-detect observation perturbations, leading to a policy taking sub-optimal or even harmful decisions with high confidence. To address this threat, it is essential to develop a framework that can assure the safety of decision-support systems prior to deployment, through both probing potential vulnerabilities and offering operators insights into the learned behavior.

In this paper, we explore methods to develop optimally timed and targeted attacks, as well as measure the attack impact and transferability within a classic reinforcement learning (RL) setting. Our methodology involves collecting attack data, designing attack strategies that produce realistic and feasible perturbations, and measuring the impact of these attacks on various properties of the RL environment. We employ a custom-built strategic game, CyberStrike, as our experimental environment to validate our framework and visualizations. 

Our contributions are threefold: First, we develop an analysis and visualization framework to help operators and researchers understand a policy's learned behavior and vulnerabilities. Second, we develop a method to programmatically discover and rank the property impacts of attacking various observation indexes at various steps of an episode. Third, we test the transferability of adversarial attacks across agents trained with different algorithms and learning curricula. 

\section{Related Work}

Conducting adversarial attacks on neural network policies is not as groundbreaking of a concept now as it was when first explored in \cite{huang2017adversarialattacksneuralnetwork}, which extended previous work in adversarial attacks in the computer vision domain such as Fast Gradient Sign Method (FGSM) \cite{fgsmgoodfellow} and Carlini-Wagner attacks \cite{carliniwagner}. Though, solely researching adversarial attacks on action selection may be too shallow of a target to propagate meaningful influence towards a desired environmental property outcome. Methods introduced in \cite{LTL_DRL} and \cite{AGTS} suggest utilizing formal language, such as Linear Temporal Logic (LTL) to define objectives and constraints for DRL policies, assist in reward function design and explain behavior patterns of policies acting within a Markov Decision Process (MDP). More recent work \cite{modelchecking} explores adversarial methods to impact atomic properties of the formalized environment; employing the aforementioned action-influencing adversarial attacks as building blocks to influence higher-level properties of the environment MDP and LTL objective specification. In addition to considering the formal logic definitions of a policy and environment when formulating attacks, we also build upon analysis and visualization techniques utilizing the internal learned models of a policy, or the Semi-Aggregate Markov Decision Processes (SAMDP) \cite{SAMDPpaper}.  SAMDP analysis first aggregates observed agent behavior into meta-data sets, then clusters model-activation layer embeddings within a two-dimensional space, and finally visualizes the behavioral patterns within this embedding space with respect to atomic properties of interest. SAMDP's are used by \cite{arlin} to characterize policies and their vulnerabilities, and we supplement these methods by illustrating the impact of adversarial attacks on environment properties at various regions of the activation embedding space. While DRL algorithms train policies to act within an environment MDP, the policy's empirical action patterns within the environment are a proxy representation of some subset of the environment MDP itself; suggesting that the identification of vulnerable embedding-space regions and observation indexes of one policy may transfer to other policies acting within the same environment \cite{transferability, transferability2}, even if the policies were not trained with the same algorithm.

We build upon this research to develop an analytical framework to determine the optimal attack timing, attack targets, and observation perturbations to deliberately impact environment properties of interest and visualize this impact.

%https://arxiv.org/pdf/2212.05337.pdf << 
% Talk about SAMDP used to visualize/explain policy 
% Talk about PIA paper 
% Talk about adversarial attacks / attack transfers generally

%\newpage
% TODO: DELETE THIS \newpage LATER!!

\section{Methodology}

\subsection{Collecting Attack Data}

\begin{figure}[h!]
    \centering
    \includegraphics[scale=0.6]{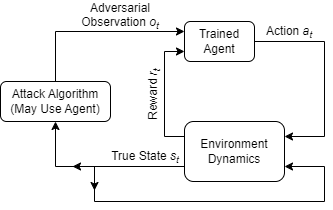}
     \caption{RL interaction loop with an attack injected at time step $t$. This time step ends with the environment dynamics using the agent's action $a_t$ and the true state $s_t$ to compute the next state $s_{t+1}$ and the reward $r_{t+1}$. Time step $t + 1$ may or may not have an attack.}
    \label{fig:attack_loop}
\end{figure}

The process for injecting  adversarial attacks into the classical Reinforcement Learning (RL) loop is shown in Figure \ref{fig:attack_loop}. Importantly, instead of directly altering the underlying state variable $s_t$ that influences the next step of the environment dynamics, our adversary is only allowed to change \textit{the agent's perception of} $s_t$ by sending a perturbed adversarial observation $o_t$ to the agent. As such, the adversary can only influence environment dynamics indirectly via the agent -- the attacks engineer $o_t$ in an attempt to control or alter the agent's action $a_t$.

\begin{figure}[h!]
    \centering
    \includegraphics[scale=0.5]{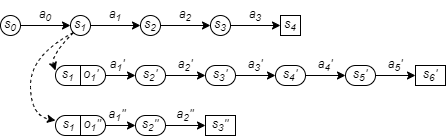}
    \caption{Example set of attacked episode simulations stemming from an unattacked episode with $4$ actions (top line). In this scenario, the attack algorithm ran several attacks on state $s_1$ (at time step $1$), and two of these attacks induced adversarial actions $a_1'$ and $a_1''$ that sufficiently differ from the original action $a_1$, meeting the criteria for simulating the rest of the episode. Taking the adversarial actions $a_1'$ and $a_1''$ from state $s_1$ will produce states $s_2'$ and $s_2''$, respectively, which may or may not differ from $s_2$.}
    \label{fig:attack_sim}
\end{figure}
%  bb: "The attacks engineer o_t..." ----> May sound better as "The attacks pertub o_t"

%[TODO: Summarize the attacked dataset collection process, keeping in mind that Figure \ref{fig:attack_sim} already gives an example and illustrates several key points.]   
%(bb) Done?

%[TODO: Include the simple measure we currently use for testing successful ``adversarial'' actions in discrete-action environments, and explain how this test would need to change for actions with continuous components.]

% (bb) too early to mention property impacts here? I think we will introduce it in related works; but we need to up-front specify the MOTIVATION behind doing these "simulated rollouts". 
% (kg) ^^ added an intro sentence giving some motivation

To study the effects of adversarial attacks, we first obtain \textit{simulated rollouts}, depicted in Figure \ref{fig:attack_sim}. Given a state $s_t$ and deterministic action $a_t$ taken by the agent in state $s_t$, we call an adversarial attack on $s_t$ \textit{sufficiently adversarial} if the agent's adversarial observation $o_t$ makes the agent take an action $a_i'$ that sufficiently differs from $a_t$, according to some predefined distance metric over the action space and some pre-selected distance threshold. In an environment with discrete actions, for example, a sufficiently adversarial action would be a simple inequality, some action $a_t' \neq a_t$. However in environments with continuous action components, we must define \textit{sufficiently adversarial} thresholds to compare $a_t'$ and $a_t$. A simulated rollout from an attacked state $s_t$ is only carried out if the adversarial action $a_t'$ is sufficiently adversarial, so that data may be collected on the end-of-episode properties and compared to those of the unattacked trajectory, in an attempt to gauge the impact of the adversarial action. Figure \ref{fig:attack_sim} illustrates a hypothetical example of the simulated rollout process from a single state of one observed episode; 
%these simulations carried out over the sufficiently adversarial attacks on the state at time step $1$ of a given episode. This figure only illustrates the simulation process from one state of one specific episode; 
however the full simulated rollout process ranges over all attacks performed on all non-terminal states of each episode in the collected data set of agent experiences.\\
 
\noindent\subsubsection{Computational Costs and Sampling} In practice, a large proportion of all attempted attacks may be sufficiently adversarial, in which case running simulations to determine the impact of every sufficiently adversarial attack is computationally expensive. Specifically, under the ``best-case'' assumption that each environment step runs in $\Theta(1)$ time, the expected time complexity of running all of these simulated rollouts is $\Theta(LN)$, where $N$ is the number of sufficiently adversarial attacks over the whole data set and $L$ is the expected length (number of time-steps) from the attack point to the end of the episode. The experiments in this paper only explore the impacts of $N$ single attack points rather than chains of multiple attacks, which would exponentially increase the time complexity. In many cases, $L$ scales with the expected length of a full episode, often linearly.  Therefore, for environments that tend to require a large number of time-steps per episode, we can expect simulated rollouts from attacks to be expensive, particularly for those attacks that stem from early states in an episode. To combat these computational costs, stratified sampling was implemented to prune the set of attacks from which to simulate while still guaranteeing sufficient representation from desired sub-populations.

%[TODO: Do we need to discuss specific methods for stratified sampling? Or is it okay to spare such details?]
%[TODO (bb): I'd say we only need to include if we are trying to fill space after rest of content.]
% https://arxiv.org/pdf/2212.05337.pdf      <<< Link to PIA paper that we should reference occasionally
% i.e. We employed an FGSM for our adversarial attack algorithm as suggested in []  

%
\subsection{Attack Strategy Design}

An \textit{attack strategy} is an algorithm that decides how and when to attack the agent. In an effort to select attack strategies that produce ``realistic'' attacks, we propose the following rough criteria for assessing attack realism:
\begin{itemize}
    \item \textbf{Feasibility:} For an attack at time $t$, adversarial observation $o_t$ must lie in the environment's state space.
    \item \textbf{Realistic Perturbation:} For an attack at time $t$, the perturbation of $s_t$ should be restricted to a known (and ideally small) subset of components of the state vector, such that this perturbation could realistically model a sensor inaccuracy or malfunction in a real-world implementation of the RL environment.
    \item \textbf{Low Severity:} Across all time-steps in a given episode, the average ``attack severity'' (a rough measure of the attack's impact on the \textit{expected action} and next state) should be low; roughly speaking, an \textit{expected action} is one that an expert human operator would take if they were the agent. In other words, attacks should be sparse with respect to time, especially those known to have severe impact on the expected action. ``Benign'' perturbations (those that ought to have little or no impact on the expected action) may be performed more frequently but will be filtered out of the simulated rollout process if the induced agent action is not sufficiently adversarial.
\end{itemize}

The attack strategies in our experiments satisfy the second and third bulleted conditions by limiting each perturbation to a single state component and limiting each simulated rollout to one attack (equivalently, after beginning a simulated rollout from a sufficiently adversarial attack, do not attack further). Still, this simple attack method must use environment context to guarantee that the first bulleted condition holds. In general, additional environment context will be necessary to measure attack severity and to design more sophisticated, multi-index perturbation attack strategies. In the Section \ref{sec:experimental-setup}, we illustrate an example of benign perturbations in a specific RL environment.

In addition to constraining attacks to certain time-steps and certain state components, an attack strategy involves a \textit{perturbation algorithm} that specifies a way to perturb the state vector within realistic bounds. Our experiments are limited to \textit{targeted attacks}, whose perturbation algorithms deliberately alter the observation in a way that encourages the agent to take a specific adversarial action $a_\text{adv}$. Whether such an attack ends up being \textit{sufficiently adversarial} only depends on the normal action $a$ and the attack-induced action $a'$, with no additional dependence on $a_\text{adv}$. However, our framework allows attack strategies to employ any perturbation algorithm, targeted or untargeted, as long as the three bulleted realism criteria are met.

Among the adversarial perturbation algorithms, the CW and FGSM attacks are particularly notable. CW attacks \cite{carliniwagner}, are optimization-based methods that generate minimal perturbations capable of misleading models. Conversely, FGSM \cite{fgsmgoodfellow}, is a gradient-based attack that quickly creates adversarial examples by leveraging the gradients of the loss function with respect to the input data. We default to using FGSM for our experiments, although the experimental framework is agnostic to the perturbation algorithm used.

% NOTE (KG): changed "attack algorithm" to "perturbation algorithm" so the language would conflict less with "attack strategy"
%[TODO: Discuss the other side of the ``attack algorithm'', (FGSM, etc) which targets a specific action and uses gradients to estimate the best perturbation that lies within the state space]
%[TODO: Reference FGSM/CW and discuss mathy stuff here.]

\subsection{Measuring Attack Impact}
% should probably reference the PIA model checking paper. Here/related-works or both.
\noindent\subsubsection{Defining a Property} Attack impact is measured with respect to a handful of \textit{properties} of interest that are chosen in advance. Each property captures certain information about the agent's experience in the environment up until the point at which the property is measured; as such, a property value attributed to some time step of an episode should only depend on environment variables (observed and/or latent) and actions that were realized at or before that time step. All properties of interest should be able to be computed at the very end of each episode. Certain properties, such as win/loss outcome, may \textit{only} be known at the very end of the episode; however, other properties, such as number of prior steps that incurred some kind of environment-based reward penalty, can be computed at any step during the episode.
% "After a suite of properties of interest ***has*** been selected" ... or is it "HAVE" ???
% (KG): should be "has" because the "suite of properties" is the object of the verb "select". The "suite" is a singular collection. More details: https://english.stackexchange.com/questions/479493/a-set-of-is-or-a-set-of-are#:~:text=If%20it's%20desirable%20that%20everyone,use%20the%20plural%20%22are%22.
% (KG): I chose to use "suite of properties" instead of just "properties" since I think it conveys better how the properties are not merely individual numbers being grabbed, but rather *functions* we are interested in running on the simulated rollout.
After a suite of properties of interest have been selected, these properties are logged during all data collection rollouts for both attacked and unattacked episodes. These logged properties, particularly those at the end of each episode, are used for downstream ``property impact analysis" \cite{modelchecking}.\\ % TODO: reference paper

\noindent\subsubsection{Mathematically Modeling Properties} To describe so-called ``property impact'' from an attack at time step $t$, we start by modeling the end-of-episode value of each $i$-th property $P_i$ in our suite as a random variable $P_i(s_t, I_t, o_t)$ that is a function of three arguments:
\begin{itemize}
    \item $s_t$ is the state vector at time step $t$;
    \item $I_t$ is a collection of other hidden environment information (including property logs) at time step $t$; and 
    \item $o_t$ is the (potentially adversarial) observation sent to the agent at time step $t$ (when no attack is present, one has $o_t = s_t$).
\end{itemize}
When $P_i$ can be expressed meaningfully as a scalar value, expected value of the given property mode becomes a relevant measure. One may estimate $\mathbb{E}(P_i(s_t, I_t, o_t))$ by running repeated trials of simulations from the same $(s_t, I_t, o_t)$ and uniformly averaging the observed values of $P_i(s_t, I_t, o_t)$. %; call this estimate $\bar{P_i}(s_t, I_t, o_t)$.
Given an attack at time step $t$ that replaces the true state $s_t$ with an adversarial observation $o_t$, the \textit{attacked value} of property $P_i$ is defined to be $P_i(s_t, I_t, o_t)$, and the \textit{unattacked value} of property $P_i$ is defined to be $P_i(s_t, I_t, s_t)$ (in the latter, the agent's observation matches the true state).\\

\noindent\subsubsection{Impact Metrics} To measure the impact of the given attack at time step $t$ on property $P_i$, we feed the attacked and unattacked values of $P_i$ into an \textit{impact metric} function $D(\cdot\, , \,\cdot)$ as the first and second arguments, respectively. Note that the resulting impact value $D(P_i(s_t, I_t, o_t), P_i(s_t, I_t, s_t))$ is a random variable. One simple impact metric for a scalar-valued property $P_i$ is the difference $P_i(s_t, I_t, o_t) - P_i(s_t, I_t, s_t)$, which conveys both magnitude and direction of the observed change in the property induced by the attack at time step $t$. Another simple impact metric for \textit{any} property $P_i$ is
$$\begin{cases} 1 & \text{if } P_i(s_t, I_t, o_t) \neq P_i(s_t, I_t, s_t) \\ 0 & \text{if } P_i(s_t, I_t, o_t) = P_i(s_t, I_t, s_t). \end{cases}$$
While well-defined, this second impact metric may lose saliency when $P_i$ has any component that ranges over a continuous domain. To illustrate one way to combat this issue, if the property $P_i$ resides in some metric space with a distance metric $d(\cdot\, , \,\cdot)$, then one could construct an impact metric such as
$$\begin{cases} 1 & \text{if } d(P_i(s_t, I_t, o_t), P_i(s_t, I_t, s_t)) > d^* \\ 0 & \text{if } d(P_i(s_t, I_t, o_t), P_i(s_t, I_t, s_t)) \leq d^* \end{cases}$$
for some distance threshold $d^*$. For each impact metric function $D$ that is invoked on a given property $P_i$ and a given attack $s_t \rightarrow o_t$, one can estimate $\mathbb{E}(D(P_i(s_t, I_t, o_t), P_i(s_t, I_t, s_t)))$ using repeated trials. Specifically, if we let $\mathcal{P}_i'$ be the list of all observed values of $P_i(s_t, I_t, o_t)$ and $\mathcal{P}_i$ be the list of all observed values of $P_i(s_t, I_t, s_t)$, then $\mathbb{E}(D(P_i(s_t, I_t, o_t), P_i(s_t, I_t, s_t)))$ is estimated by the uniform average
$$\frac{1}{|\mathcal{P}_i'||\mathcal{P}_i|}\sum_{p_i' \in \mathcal{P}_i'}\sum_{p_i \in \mathcal{P}_i} D(p_i', p_i),$$
where both summations range over all elements, including repeats, in the lists $\mathcal{P}_i'$ and $\mathcal{P}_i$.

%For whichever property impact metrics are computed for a given property $P_i$ and a given attack, the impact values are averaged uniformly across all pairs of realizations of the random variables $P_i(s_t, I_t, o_t)$ and $P_i(s_t, I_t, s_t)$ resulting from the repeated trials of attacked and unattacked rollouts, respectively, from time step $t$ [TODO: state this more clearly!!][(bb): agree]
%The selected property impact suite's "impact values" are computed for a given property $P_i$ and a given attack.  averaged uniformly across all pairs of simulated random variables

\section{Experiments}
The motivation behind our experiments is threefold; First to develop an analysis and visualization framework to supplement the researcher's understanding of a policy's learned behavior and vulnerabilities, second to analyze the property impact of attacking various observation indexes, and third to test the transferability of adversarial attacks across agents of various training algorithms and learning curricula.  

%[TODO: intro into experiments? ]open to suggestions on how to break this section up, either consolidating our various experiments (property impact, attack transferability, explainability results) into one set up section and one results section, or breaking them up into their own individual sections (similar to how the causality paper separates each of their causal experiments into unique sections)

% Could move this environment explanation into the appendix if necessary. 

\begin{figure}[h!]
    \centering
    \includegraphics[scale=0.5]{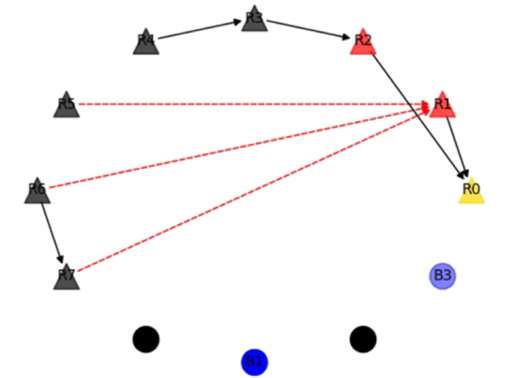}
     \caption{A notional CyberStrike state. The blue agent controls nodes B0, B1, B2, B3. Blue chooses actions which control each blue node simultaneously, locating and disabling the target red node by peeling back the layers of the red defense network until the target node is undefended. In this example, the target node ($R0$) is defended by $R1$ and $R2$. $R2$ is defended by $R3$, which is defended by $R4$. $R1$ is defended by $R5$, $R6$, and $R7$. $R6$ is also defended by $R7$. Dashed lines denote a connection marked as unknown in the agent's observation, whereas solid lines represent a known connection. The agent begins with a fully unknown network, and must use its hackers to discover the network topology enough to reveal the target node's ($R0$'s) defenders and eventually hack into the target node.}
    \label{fig:cyber-strike}
\end{figure}
\subsection{Experimental Environment}
Deep reinforcement learning is increasingly being used to discover adversarial tactics, techniques, and procedures (TTPs) within the cybersecurity domain \cite{farland}. The gym environment used for notional experiments is CyberStrike; a custom-built, strategic network-defense game wherein an agent controlling blue nodes must determine information about the red network's tree structure, and then hack into each red node's parent node recursively until reaching the target node. The CyberStrike environment is ripe for emergent, explainable learned strategy; contrasting typical control-focused benchmarks such as LunarLander-v2 or Cartpole \cite{openAIgym}. The action space is multi-discrete, made up of four blue ``hackers" that can simultaneously ``hack" or ``eavesdrop" on a collection of red nodes. If a blue hacker attempts to hack a defended red node, the red defender will counter and the blue hacker will be unavailable for the rest of the episode. The ``eavesdrop" action is only available to one of the blue hackers ($B3$), and allows the agent to stealthily learn the defenders of a red node without risking a counter from red. An example network structure from a mid-episode observation is displayed in Figure \ref{fig:cyber-strike}.  

\begin{figure}[h!]
    \centering
    \includegraphics[scale=0.55]{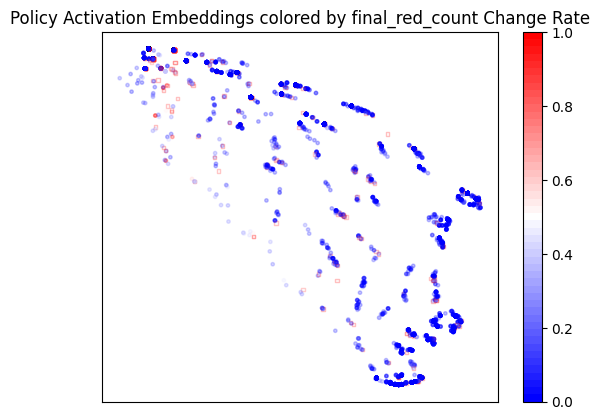}
     \caption{ This latent space representation maps a policy's CyberStrike observations from initial time-steps in the northwest region to the final time-steps in the southeast region, with an aggregation of various intermediate trajectories connecting the initial and final observations. Attacks within the denser, bluer northeast region of the space are unlikely to yield nonzero changes in final red counts, whereas attacks in the sparser and redder western regions are more likely to be successful (increase final red counts). The sparsity of activation embeddings in the western region of the latent space representation suggests the policy is less likely to have trained on observations in this region and thus is more vulnerable to adversarial attacks when acting within this region.}
    \label{fig:SAMDP}
\end{figure}

\subsection{Experimental Setup}\label{sec:experimental-setup}
First, we train a suite of both Advantage Actor Critic (A2c) \cite{a2cpaper} and Deep Q-Network (DQN) agents \cite{dqnpaper} within the CyberStrike environment. Following training we collect 10,000 state-action-metadata tuples from the frozen policies acting within CyberStrike, collecting metadata such as a policy's hidden-layer activations, observation saliency, and step-wise environment properties. Due to the absence of $\epsilon$-greedy or distribution sampling for exploration, we force a small percentage (5\%) of random actions during the data collection to inform potential adversarial targets, though the researcher may vary the percentage of random action selected depending on the environment MDP and frozen policy optimality. These collected data sets from the trained policies help to represent empirical policy behavior through activation clustering, SAMDP transition visualization, and other custom metadata visualizations. For example, Figure \ref{fig:SAMDP} shows how one can track the average change rate of a property at any attacked observation. For each collected observation, the policy's final latent activation layer is embedded in two dimensions and colored with a gradient across the aggregate change rates of a property of interest, the change rate being determined by the difference in the property value for the unperturbed versus perturbed observations. These behavioral visualizations help the researcher get a birds-eye view of policy trajectories as they relate to environment properties; while also highlighting feasible, optimally-timed, and low-severity attacks on the policy's learned strategy. We can also run adversarial attacks on (and simulated rollouts from) the observations collected in these data sets, which will be necessary for both Property Impact and Attack Transferability Analysis.\\

% Not using taxi; but maybe discuss anyway? [TODO: Discuss an example of these ``more sophisticated'' attack algorithms for the taxi environment that leverages certain benign perturbations in this environment (e.g., changing the passenger destination willy-nilly before the passenger has been picked up).]

% [TODO/QUESTION: does the observed defense network allow for defense values of ``True'' between two red assets if either of them is already destroyed?] bb answer: YES this is allowed

\noindent\subsubsection{Benign Perturbations in CyberStrike} 
In the Cyber Strike environment, one kind of benign perturbation would consist of selecting an ordered pair $(A, B)$ of distinct red assets, with at least one already compromised by a blue hack, and changing the agent's perception of whether or not $A$ defends $B$. 
Such a perturbation on the defense $A \rightarrow B$ is benign because an expert human hacker would not target $B$ if it were already compromised; and if $A$ was already compromised, then $A$ would not be able to counterattack a blue node following its hack on $B$, making the defense value for $A \rightarrow B$ irrelevant to decision-making. Therefore, our attack-discovery framework would permit this kind of benign attack to be made frequently in a single episode, since an ideal policy ought to not behave differently from any of these attacks.

% TODO: figures of SAMDP clustering / saliency?
\begin{figure}[h!]
    \centering
    \includegraphics[scale=0.5]{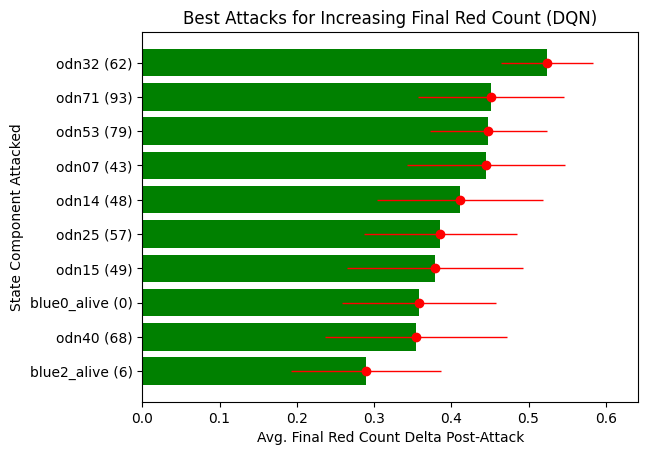}
     \caption{The Average \textit{Final Red Count} delta post-attack is aggregated per observation index, across all time-steps. Eight out of the ten most impactful attacked observation indexes are observed defense network nodes, suggesting that an attacker's best chance of increasing the final red count is to perturb the DQN agent's perception of the network structure at various adjacency nodes.}
    \label{fig:pia_finalRedCount}
\end{figure}

\begin{table*}[h!]
\centering
\includegraphics[scale=0.9]{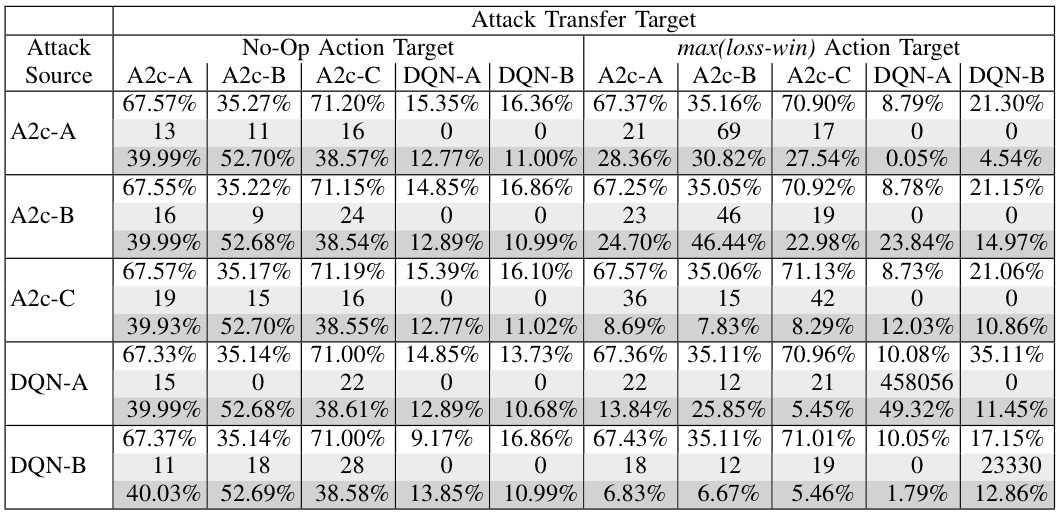}
\caption{Attack transferability results from experiments outlined in Section \ref{sec:transferability-setup}. The attacks are configured using the policy in the Attack Source column, targeting the No-Op (left) and \textit{max(loss-win)} (right) action targets. The attacks are then run on the attack-source policy and transferred to the other four policies of interest. Three metrics are recorded per cell: transferability success rate (white sub-cell), target-transferability count out of one million (light gray sub-cell), and sub-action target-transferability success proportion (dark gray sub-cell). \textit{Self-attacks}, where the attack source and target policy are the same, are also included in this table. }
\label{table:1}
\end{table*}

\begin{figure}[h!]
    \centering
    \includegraphics[scale=0.5]{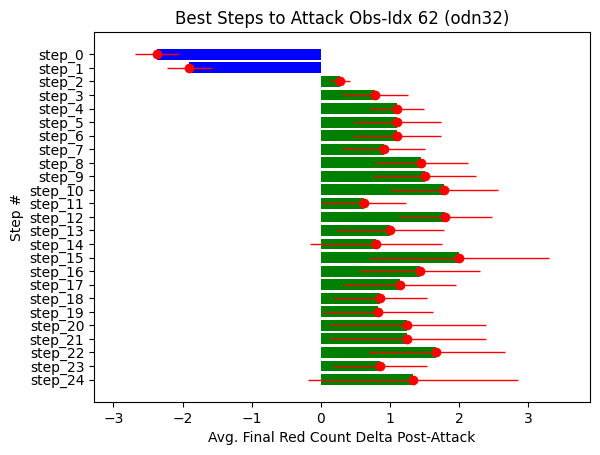}
     \caption{The \textit{Final Red Count} delta is plotted at each step for all attacks on $odn32$'s value in the DQN's observation. This plots suggests that attacking $odn32$ at the first two steps may have negative effects for the attacker, whereas attacks from step 2 onward correspond with an increase in red nodes (thus a decrease in blue win percentage) compared to an unattacked trajectory. }
    \label{fig:pia_step_finalRedCount}
\end{figure}

\subsection{Property Impact Attack Analysis}
In order to measure and compare the aggregate property impact of attacking various observation indexes, we must run adversarial attacks perturbing each observation index for each collected observation tuple. Simulated rollouts are performed only from adversarial attacks inducing an action $a_i'$ that is sufficiently different from the original action $a_i$. Environment properties are measured at the terminal state of the simulated rollout and compared to environment properties of the unattacked trajectory, as to gauge the property impact of a given attack. We can aggregate these impact metrics across step numbers and observation indexes to answer questions about the ideal time-step or observation index to attack with respect to some environment property the attacker wishes to impact. In CyberStrike, we measure the attack impact on properties such as win percentage, final red count, final blue count, and trajectory length. Figure \ref{fig:pia_finalRedCount} displays a ranked aggregation of final red count deltas across attacked observation indexes for a DQN policy. The policy's most impactful observation index attack with respect to final red count is a perturbation of the value of the observed defense network node 3:2 (\textit{odn32}), denoting if $R3$ is defending $R2$ or if this connection is unknown. In the notional example in Figure \ref{fig:cyber-strike}, $R3$ \textit{is} defending $R2$, so a perturbation obscuring this information may cause the agent to take an action hacking the defended $R2$ node, whereas an optimal decision would be to hack $R2$'s defender node, $R3$, first. Figure \ref{fig:pia_step_finalRedCount} displays the impact aggregation across time-steps for all attacks on the \textit{odn32} observation value, suggesting that attacks early in an episode, time-steps 0 and 1, may have negative consequences for the attacker by decreasing their average final red count. Figure \ref{fig:pia_step_finalRedCount} shows that time-steps 10 and 15 lead to the largest average final red count increase, suggesting an adversary may have the greatest impact on final red count in the middle of an episode, rather than at the beginning or end. 

\subsubsection{Property Impact Analysis Results}
The Property Impact Analysis results indicate that adversarial attacks may exert both positive and negative influences on the attacker's desired outcome, dependent on the time-step when the attack is brokered. 
By strategically timing the manipulation of the most vulnerable observation components of a policy, we are able to observe significant variations in policy behavior, leading to notable changes in the environment properties and game outcome; thus demonstrating the ability to deliberately impact an external environment property by choosing a specific adversarial attack target at a specific time-step. 

Figure \ref{fig:pia_step_finalRedCount} displays the ``final red counts" property outcomes when a red attacker attacks the \textit{odn32} observation component at various time-steps. Specifically, we find that attacking the observation at the initial steps of the game may lead to an unexpected decrease in the final red count, which runs contrary to the attacker's objective. However, as the game progresses, attacks on specific observations can result in more favorable increase in the final red count, aligning with the attacker's strategic intentions. This finding underscores the dynamic nature of learned policies, even within simple environments. It highlights the delicate interplay between attack targets, how those action targets influence future behavior, and how that future behavior affects the environment properties and the ultimate objective outcome.

% BB TODO:ask Kyle; how does action selection work for a2c? Were we takign argmax of policy distribution or sampling from dist? It changes how I interpret these results. 

\subsection{Attack Transferability Analysis}
In addition to analyzing the property impact of various adversarial attacks on a single policy, we also analyze the transferability of an attack trained with one policy and deployed during another policy's execution. 
\subsubsection{Transferability Metrics}
In order for an attack to be \textit{transferable}, the attack (parameterized by policy $\pi_i$) must induce a \textit{sufficiently adversarial} action when deployed on some other policy $\pi_j$. In CyberStrike, an attack parameterized by a policy $\pi_i$ is counted as \textit{transferable} if it induces an action different from the action taken by a policy $\pi_j$ from the unattacked observation. A targeted attack, parameterized by $\pi_i$, is counted as \textit{target-transferable} if it induces the attack's \textit{target action} on the new policy $\pi_j$. Due to the multi-discrete nature of CyberStrike's action space (which has four sub-actions), we can also measure the proportion of the induced sub-actions matching the target sub-actions, or the \textit{sub-action target-transferability}. 
\subsubsection{Transferability Experimental Setup}\label{sec:transferability-setup}
We employ Automated Domain Randomization (ADR) and Curriculum Learning (CL) across the action and counter-action effectiveness dimensions, coined action stickiness by \cite{stickyactions}, to increase the variation in learned strategies, providing more heterogeneous policy targets for attack-transfer. We will analyze attack transferability across a suite of five policies: A2c-ADR+CL (A), A2c-ADR (B), A2c-CL (C), DQN-CL (A), DQN-deterministic (B)). 

Training curriculum and hyperparameter details for the policies are available in the appendix. For each policy, we use two action-targets for transferability analysis: the 0-action (No-op) and the \textit{max(loss-win)} action. The \textit{max(loss-win)} action is computed by counting each collected action's usage within winning and losing trajectories; if  the action was used $U_L$ times in losing trajectories and $U_W$ times in winning trajectories then the \textit{(loss-win)} value is $U_L - U_W$, and the \textit{max(loss-win)} action target maximizes this value.  

After determining the \textit{max(loss-win)} action-target for each source policy, we run the transferred adversarial attacks. For each observation in each target policy's collected dataset, we run an adversarial attack for each action target and collect metrics regarding \textit{transferability}, \textit{target-transferability}, and \textit{subaction-target-transferability}. We hypothesize that attacks may be more transferable between policies of the same DRL algorithm (A2c-X $\rightarrow$ A2c-Y, or DQN-X $\rightarrow$ DQN-Y), however the target policy should be the biggest factor in transferability, regardless of target action or source policy. We also hypothesize that the \textit{max(loss-win)} action target may be more easily induced compared to the No-Op action, because the No-Op action should not be taken by an optimal or near-optimal policy, whereas the \textit{max(loss-win)} actions are empirically taken by the source policies during losses.

\subsubsection{Transferability Results}
The policy target is indeed the greatest factor on transferability success rates, especially for the A2c policies where we see roughly the same transferability success rates per policy target, across all attack sources and action targets. It is also worth noting that the \textit{max(loss-win)} action target induces target-transfers most often, but still sparsely, for A2c policies. DQN self-attacks induce the target action 45.8\% (DQN-A) and 2.3\% (DQN-B) of the time, however attacks transferred to DQN policies never induce the target action. Contrarily, A2c self-attacks induce the target action at roughly the same rate as attacks transferred to A2c policies. The variation in sub-action target-transferability per-row in the \textit{max(loss-win)} block can be attributed to the \textit{max(loss-win)} action being different for each source policy.
\section{Discussion \& Conclusions}
 The results suggest the ability to influence agent behavior, and thus future environment properties, is controllable through optimally timed, deliberately chosen observation perturbations. This capability, paired with the result showcasing varying levels of attack transferability across algorithm types, highlights the urgent need for robust defense mechanisms and adversarial evaluation schemes to safeguard decision-making policies from the threat of adversarial influence, especially in high-stakes environments. The results also suggest that policies trained with some algorithms, like A2c, may be more vulnerable to transferred attacks than others, such as DQN in this specific experimental setting; and transferability must be measured on a per-algorithm basis. The presence of observation-dependent and time-dependent vulnerabilities implies the existence of training and fine-tuning methods to guard against these vulnerabilities, though we have not explored methods to do so in this paper and leave that to future research. 
 
 While this paper focuses on using adversarial attacks to probe and analyze the behavior of policies trained through DRL algorithms, the same behavioral analysis may be conducted on LLM-based agentic architectures, albeit with language-based attacks and alternate metadata for t-SNE embeddings. We will leave this to future adversarial analysis research.
 
\section*{Impact Statement}
The paper presents work whose goal is to advance the field of machine learning, specifically regarding deep reinforcement learning explainability and adversarial analysis. As society continues to adopt DRL and AI solutions broadly, explainability and evaluation methods such as those presented in this paper will help provide frameworks to assure and gain trust of these systems. 
\section*{Acknowledgments}
The authors thank Guido Zarrella and Dr. Chris Niessen for their advisory roles throughout the research and development process. This work was funded by the 2023 MITRE Independent Research and Development Program.

\bibliographystyle{icml2025}
\bibliography{references}

\newpage
\newpage
\newpage

\appendix
\onecolumn

\section{Appendix}

\subsection{Publicly Available Code}

Code for the Cyberstrike environment, DRL-SAT analysis repository, and training repository will be open sourced at: https://github.com/mitre/drlsat. 

\subsection{CyberStrike}

Cyberstrike is a highly customizable network defense environment, initialized with the following configuration parameters. The values listed were used for experiments, with the exception of standard deviations of ADR variables: 

\begin{lstlisting}[label=scenario.yml,caption=CyberStrike Configuration File]
adr_variables: 
  - id: adr_0v1
    type: adr_normal_range
    parameters:
      mean: 1.0
       # standard deviation varies 
       # with ADR & CL
      stdev: 1.0
      maximum: 1.0
      minimum: 0.1

  - id: adr_0v2
    type: adr_normal_range
    parameters:
      mean: 1.0 
      stdev: 1.0
      maximum: 1.0
      minimum: 0.1

  - id: adr_1v0
    type: adr_normal_range
    parameters:
      mean: 1.0
      stdev: 1.0 
      maximum: 1.0
      minimum: 0.1

  - id: adr_2v0
    type: adr_normal_range
    parameters:
      mean: 1.0
      stdev: 1.0 
      maximum: 1.0 
      minimum: 0.1
scenario:
  red:
    assets:
      - is_target: true #0
        type: 0
        is_alive: True
      - is_target: false #1
        type: 0
        is_alive: True
      - is_target: false #2
        type: 0
        is_alive: True
      - is_target: false #3
        type: 0
        is_alive: True
      - is_target: false #4
        type: 0
      - is_target: false #5
        type: 0
        is_alive: True
      - is_target: false #6
        type: 0
        is_alive: True
      - is_target: false #7
        type: 0
        is_alive: True
    defense_network:    
      - [1, 2] #red node 0 defended by [1,2]
      - [ 5, 6, 7 ]# red node 1 is defended by [5, 6 and 7]
      - [3] #red node 2 defended by 3
      - [4]         
      - [] #4
      - [] #5
      - [] #6
      - [6] #red node 7 is defended by 6
  blue:
    assets:
      - type: 1
        loss_cost: 20
        use_cost: 2
      - type: 2
        loss_cost: 20
        use_cost: 2
      - type: 2
        loss_cost: 20
        use_cost: 2
        is_alive: True
      - type: 3
        loss_cost: 10
        use_cost: 5
  effect_probability: 
  # type {row_idx} effectiveness 
  # hacking type {col_idx}
      - [ 0,       adr_0v1, adr_0v2, 0]
      - [ adr_1v0, 0,       0,       0]
      - [ adr_2v0, 0,       0,       0]
      - [0,        0,       0,       0]

\end{lstlisting}

\subsection{Observation Space}

The observation space in CyberStrike consists of "$alive$" and "$type$" information for all blue assets, "$alive$" and "$type$" and "$is\_target$" information for red assets, and the observed defense network, from blue's perspective. This information is flattened into an array and passed to the agent as a flat tensor.  The size of the flat tensor is formally 
$$3*(num\_blue+num\_red) + num\_red^2$$

\subsection{Action Space}

The action space in CyberStrike is multi-discrete, with each blue asset capable of being paired to some red asset (or no red asset) for any given multi-discrete action. This means the action space linearly increases as we increase the number of red or blue assets in the configuration. Formally the action space is of size 
$$num\_blue * (num\_red + 1)$$

\subsection{Strategy and Optimality}

The optimal strategy in CyberStrike requires using eavesdrop assets to discover the defense network nodes, and then utilizing hacking assets to infiltrate the defense network, hacking undefended assets first, until the target is reached through recursive hacks. In the absence of adversarial attacks, DRL policies optimize towards this behavioral pattern. 

\begin{figure}[]
    \centering
    \includegraphics[scale=0.65]{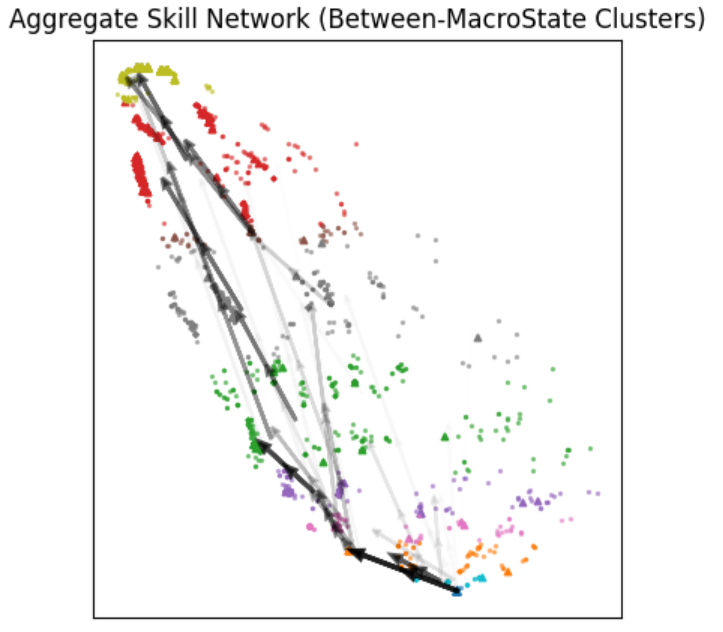}
     \caption{Embedded policy activation vectors are colored by their cluster, determined by Chinese-Whispers, and marked with aggregate skill-transition arrows. The shading of the arrows represent the empirical likelihood of the policy transitioning from one cluster to another; thus the most-travelled trajectories are marked by the darkest-shaded arrow path.}
    \label{fig:colorbycluster}
\end{figure}

\subsection{Curriculum Learning and Automated Domain Randomization}
We randomize the action effectiveness variables for Curriculum Learning (CL) and Automated Domain Randomization (ADR) policies. For the CL policies, the curriculum incrementally adds new variables to randomize as level difficulty increases. We increase the environment level whenever the learning agent reaches 90\% on its current level.  For instance, the CL agent starts training in a fully deterministic environment. Once 90\% win-rate is reached, the environment randomizes one the four effect probabilities, sampling from a truncated normal distribution centered around 1, standard deviation of 1, and minimum and maximum of 0 and 1. As the agent reaches 90\% win-rate on this second level, the environment randomizes yet another action-effectiveness dimension, until eventually all four variables are sampled with a standard deviation of 1.0 in the last level. During purely ADR training, we sample from this truncated distribution with a standard deviation of 1.0, for each of the four action effectiveness dimensions; and this pure-ADR level is identical to the final, fully-randomized level on the CL-denoted policies. The policy denoted ADR+CL (A2c-A) is trained with a curriculum that increases the standard deviation of all effectiveness probabilities by 25\% every level. Once 90\% win-rate is reached on the deterministic level 1, the agent begins training on level 2, where there is a 0.25 standard deviation for the action stickiness sampling distribution centered around 1. This ADR+CL lesson plan increases the standard deviation from $0 \rightarrow 0.25 \rightarrow 0.5 \rightarrow 0.75 \rightarrow 1.0$. 

\subsection{Training hyperparameters}
All DRL policies were trained with either DQN or A2c, utilizing the standard deep Q-learning algorithm \cite{dqnpaper} and the standard advantage actor critic algorithm introduced in \cite{a2cpaper}. For the DQN policies, we use a discount factor of .99, replay ratio of 4, target update tau of 0.05 with an interval of 250, an Adam optimizer, clip grad norm of 10, and a learning rate of $3e-4$. The $\epsilon$-greedy exploration module initializes at 1.0, decays by a factor of .99 to a minimum epsilon of 0.01.

For the A2c policies, we use a discount factor of 0.99, actor learning rate of 1.5e-4, critic learning rate of 3e-4, value loss coefficient of 0.5, entropy loss coefficient of 0.01, Adam optimizer, and clip grad norm of 10. 
The networks ingest a flat input layer of varied size, depending on the size of the CyberStrike configuration. In the CyberStrike configuration used for experiments, where we have 8 red nodes and 4 blue hackers, the input size is 100. The hidden dimension, and thus the size of the activations used for embedding and clustering, is set to 256 by default. 
Policies are trained through their curriculum, until 90\% win-rate is reached on the final level. At this point, policies are frozen, evaluated, and collected for analysis. 
Both the actor network and DQN are instantiated as follows: 

\begin{lstlisting}[label=DQN and Actor Network,caption=DQN and Actor Network]
hidden_dim = 256
self.fc = torch.nn.Sequential(
    nn.Linear(in_shape[0], hidden_dim),
    nn.ReLU(),
    nn.Linear(hidden_dim, hidden_dim),
    nn.ReLU(),
    nn.Linear(hidden_dim, out_shape[0]),
)
\end{lstlisting}

The critic network for A2c training is instantiated as follow:

\begin{lstlisting}[label=Critic Network,caption=Critic Network]
hidden_dim = 256
self.fc = torch.nn.Sequential(
    nn.Linear(in_shape[0], hidden_dim),
    nn.Tanh(),
    nn.Linear(hidden_dim, hidden_dim),
    nn.Tanh(),
    # Single output neuron for value function.
    nn.Linear(hidden_dim, 1),
)
\end{lstlisting}

\subsection{Analysis hyperparameters}

For SAMDP analysis, we utilize the policy network activation vectors to create t-Distributed Stochastic Neighbor Embeddings (t-SNE) \cite{tsne} with a perplexity of 132. We utilize Chinese-Whispers \cite{chinesewhispers} clustering algorithm with a critical distance of 15.0 to cluster the policy activation vectors, and color the associated 2d embedded points with a unique cluster color. In addition to coloring by cluster, shown in Figure \ref{fig:colorbycluster}, we can also color by adversarial or atomic property attributes as in Figure \ref{fig:SAMDP}. 

\label{sec:reference_examples}

\end{document}